\documentclass[conference]{IEEEtran}
\IEEEoverridecommandlockouts
\usepackage{cite}
\usepackage{amsmath,amssymb,amsfonts}
\usepackage{algorithmic}
\usepackage{graphicx}
\usepackage{textcomp}
\usepackage{xcolor}

\def\BibTeX{{\rm B\kern-.05em{\sc i\kern-.025em b}\kern-.08em
    T\kern-.1667em\lower.7ex\hbox{E}\kern-.125emX}}

\IEEEoverridecommandlockouts

\begin{document}

\title{Dual-Weight Particle Filter for Radar-Based Dynamic Bayesian Grid Maps\\
%\thanks{Identify applicable funding agency here. If none, delete this.}
}

\author{\IEEEauthorblockN{1\textsuperscript{st} Max Peter Ronecker}
\IEEEauthorblockA{\textit{Institute of Automation and Control} \\
\textit{Graz University of Technology}\\
Graz, Austria \\
max.ronecker@tugraz.at}
\and
\IEEEauthorblockN{2\textsuperscript{nd} Michael Stolz}
\IEEEauthorblockA{\textit{Institute of Automation and Control} \\
\textit{Graz University of Technology}\\
Graz, Austria \\
michael.stolz@tugraz.at}
\and
\IEEEauthorblockN{3\textsuperscript{rd} Daniel Watzenig}
\IEEEauthorblockA{\textit{Institute of Automation and Control} \\
\textit{Graz University of Technology}\\
Graz, Austria \\
daniel.watzenig@tugraz.at}
}

\maketitle
\IEEEpubid{\begin{minipage}{\textwidth}\ \\[12pt] \centering
  \copyright 2023 IEEE. Personal use of this material is permitted.  Permission from IEEE must be obtained for all other uses, in any current or future media, including reprinting/republishing this material for advertising or promotional purposes, creating new collective works, for resale or redistribution to servers or lists, or reuse of any copyrighted component of this work in other works.
\end{minipage}} 
\IEEEpubidadjcol
\begin{abstract}
Through constant improvements in recent years radar sensors have become a viable alternative  to lidar as the main distancing sensor of an autonomous vehicle. Although robust and with the possibility to directly measure the radial velocity, it brings it's own set of challenges, for which existing algorithms need to be adapted. One core algorithm of a perception system is dynamic occupancy grid mapping, which has traditionally relied on lidar. In this paper we present a dual-weight particle filter as an extension for a bayesian occupancy grid mapping framework to allow to operate it with radar as its main sensors. It uses two separate particle weights that are computed differently to compensate that a radial velocity measurement in many situations is not able to capture the actual velocity of an object. We evaluate the method extensively with simulated data and show the advantages over existing single weight solutions. 
\end{abstract}


\section{Introduction}

Accurately perceiving the environment is one of the key challenges when developing autonomous or highly automated vehicles \cite{b0}. A framework commonly used to represent and generate a map of the surroundings is occupancy grid mapping. Introduced in \cite{b1}, occupancy grid maps split the environment into independent cells that store information about occupancy at the respective cell position. They accumulate measurement information over time using a Bayes filter update process \cite{b2}. Initially developed assuming a static environment, dynamic objects can cause issues, in form of fragments due to not propagating occupancy of moving objects correctly. Hence, the framework has been further improved by storing dynamic information inside the cells \cite{b3} or track it using particle filters \cite{b4}. This allows a concise and combined representation of the static and dynamic environment which can be further processed by planning algorithms or Multi-Object-tracking.

Most of the applications using (dynamic) occupancy grid maps rely on lidar sensors and complement it with radar \cite{b5}. However, the high cost of lidar and the challenge of processing the large amount of data provided through the sensor hinders the mass deployment of autonomous vehicles. An alternative would be to use only radar as distancing sensors. They are widely available in automotive grade at a cheaper cost. Although noisier compared to lidar, radar works in almost any environmental condition and adds the benefits of direct radial velocity measurement. In order to make radar-only based systems more reliable and safer, this paper presents a modified version of Hybrid-Sampling Bayesian Occupancy Filter (HSBOF) \cite{b6}\cite{b7} tailored towards using radar measurements as main source of information. For this a dual-weight particle filter is used to process the radial velocity and range information in an optimal manner.
Hence, the main contribution of this paper is an adaption of the particle filter that allows to track dynamic objects in the HSBOF framework using only radar, also in situations where it would usually fail with a conventional implementation.

The outline of the paper is as follows: Section \ref{section_2} briefly discusses the current state-of-the-art for dynamic occupancy grid mapping. Section \ref{section_3} explains the used algorithm, challenges when using radar and the dual weight approach. The performance of the new approach is evaluated and compared in section~\ref{section_4}. Finally a conclusion and outlook is given in section~\ref{section_5}.

\section{Related Work} \label{section_2}
\IEEEpubidadjcol
In the original dynamic Bayesian Occupancy Filter (BOF) \cite{b3} the information is contained in a four-dimensional grid, which requires discretization of the velocity and is computational expensive. In order to tackle this a significantly more efficient version of a BOF has been proposed in \cite{b4}, combining a grid representation with a particle filter. In this framework the particle filter estimates the velocity and occupancy distribution in the grid and is used to identify dynamic cells. This method has been adopted and improved in \cite{b6}\cite{b7}. In \cite{b8} a detailed fusion approach for lidar and radar using a BOF is described. It shows that combining the two sensors improves performance of the filter. The same paper also introduces a Dempster-Shafer representation \cite{b9}, \cite{b10}, \cite{b11} which can be alternatively used to fuse the information. Further adaptions which use Dempster-Shafer as an alternative to the classical BOF \cite{b12}, combine it with Random-Finite-Set theory \cite{b13}  or incorporate object-level information and multi-object-tracking \cite{b14}, \cite{b15}, \cite{b16}. All approaches so far have been mainly lidar based, complemented with radar to increase performance. Radar-only-methods for static BOFs and freespace estimation have been proposed in \cite{b17} and \cite{b18}, by including radar-sensor-specifics, like signal amplitude and field-of-view into the standard Inverse Sensor Model. A first radar-based dynamic grid has been presented in \cite{b19}, which is an adapted version of \cite{b20} to deal with the additional noise of the radar data.   

\section{Radar-based Bayesian Dynamic Grid Maps}\label{section_3}

This section describes the relevant technical details in order to understand the used framework. It is based on \cite{b6} and determines simultaneously the state of a grid cell, to be \textit{free}, \textit{static occupied} and \textit{dynamic occupied}. It also utilizes a dualism of classical occupancy grid mapping and particle filter. Furthermore, the modifications to optimize the framework towards the use of only radar data and the reasoning behind them are explained. 

\subsection{Environment Representation}

\begin{figure}[htbp]
\centerline{\includegraphics[width=\linewidth]{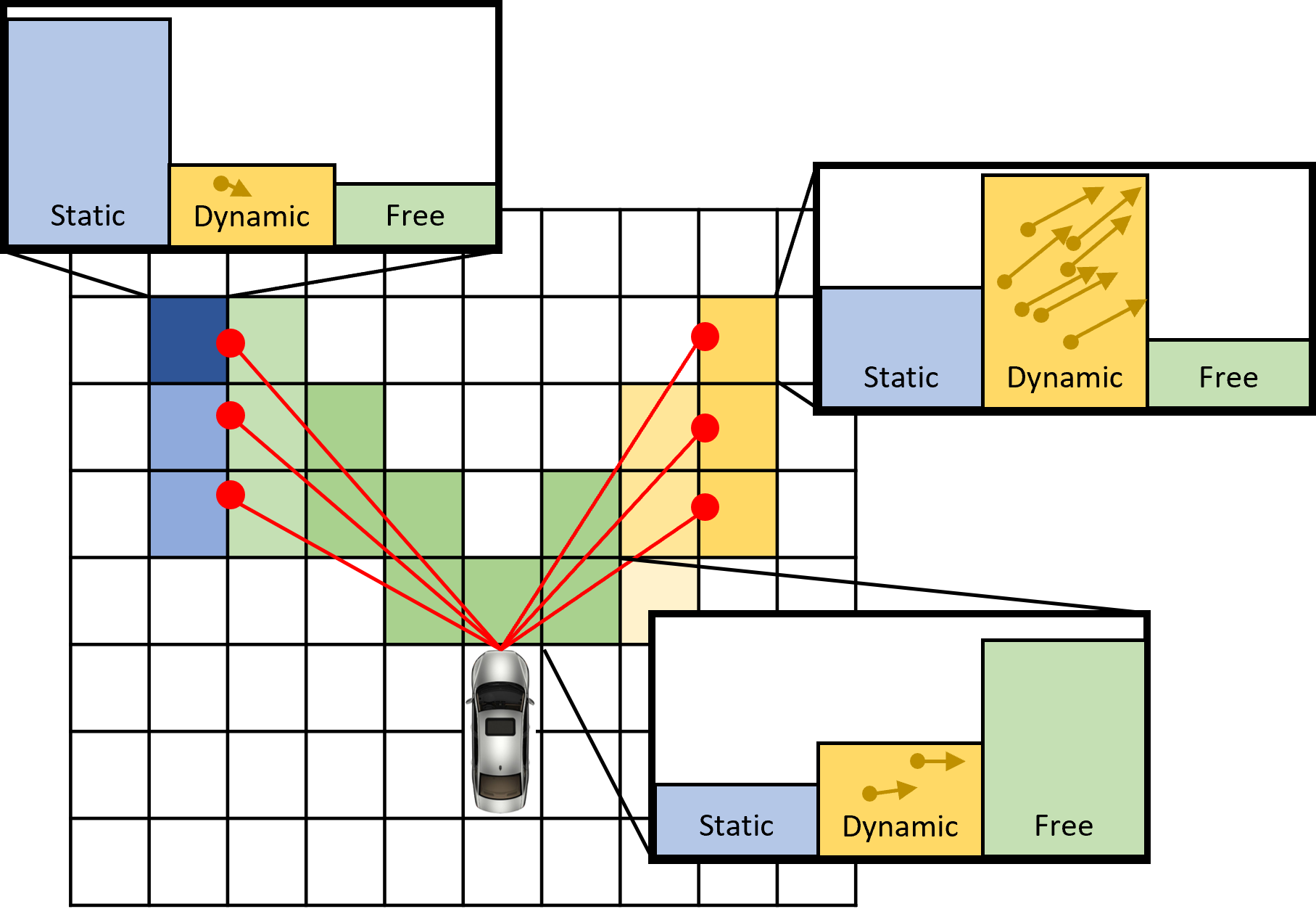}}
\caption{Each cell contains information about free space, occupancy as well as dynamics in form of particles. \cite{b6}.} 
\label{grid_environment}
\end{figure}

The surrounding environment of the vehicle is represented by a grid map, centered at the position of the ego-vehicle. 
Each cell consists of three parts, which express, if it is empty (\textit{free}), occupied by a static object (\textit{static occupied}), or occupied by a dynamic object (\textit{dynamic occupied}). 
Additionally particles with the state vector $\mathbf{x}\!=\!\left[ x, y, v_x, v_y, weight\right]^T$ are used to help in tracking the dynamic occupation. 
Each particle is assigned to a specific cell based on the position. 
Fig.~\ref{grid_environment} visualizes the concept.

\begin{itemize}
    \item \textit{Free}: Probability that the cell is empty $P(emp)$. This is identical to the free space known from classical occupancy grid mapping. Free space doesn't contain any dynamic component as empty cells are always assumed to have zero velocity
    \item \textit{Static (occupied)}: Probability of the cell to be occupied and to have velocity zero $P(occ \cap v=0)$. This slightly differs from classical occupancy grid mapping as velocity information is included when updating the occupancy.
    \item \textit{Dynamic (occupied)}: Probability of the cell to be occupied and to have non-zero velocity $P(occ \cap v\neq0)$. This component consists of a set of particles, which represent the velocity distribution of the cell. Each particle has also a weight, which expresses certainty about the similarity of the particle state to observed state given recent measurements. The sum of all particle weights represents the overall dynamic probability of a cell.  
\end{itemize}
All updates for the states are in each processing cycle first calculated independently, and secondly followed by a joint update. 
An advantage of this joint representation is that no explicit separation into static and dynamic measurements is necessary. 
This often is a challenge for radar data, hence the framework is well suited to deal with radar sensors.

\subsection{Bayesian Dynamic Grid Mapping}

\begin{figure*}[htbp]
\centerline{\includegraphics[width=\textwidth]{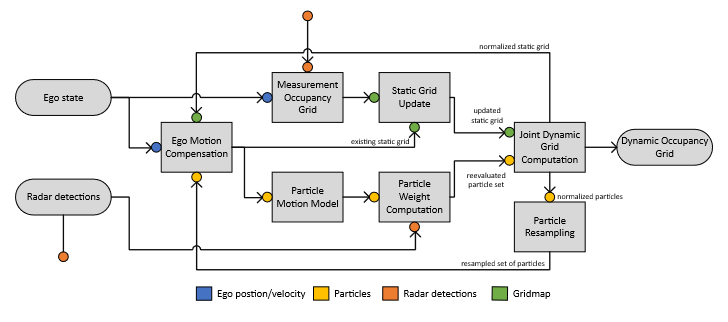}}
\caption{General structure of the used algorithm as shown in \cite{b21}. It combines classical occupancy grid mapping with a particle filter for estimating dynamics inside the grid.} 
\label{HSBOF}
\end{figure*}

In this section the relevant details for understanding the HSBOF are provided. An overview of the algorithm structure is given in figure. The focus of this section will be on how the each of the cell states are derived and the joint update process. A detailed description can be found in the original paper \cite{b6}. In Fig.~\ref{HSBOF} the general structure of the algorithm is displayed. In the following some of the key components (1-4) will be described in more detail.

\subsubsection{Radar Measurements}

A radar measurement commonly is defined by a position in polar coordinates and a radial velocity (range rate). For easier processing in the grid the measurement is transformed into Cartesian coordinates. Position is in the vehicle frame and the x-y velocity is a projection of the radial velocity into the global frame resulting in the state vector $\mathbf{x}_r$

\begin{equation}
    \mathbf{x}_r = \left[ x_r, y_r, v_{r,x}, v_{r,y} \right] ^T  \label{eq1}
\end{equation}

\subsubsection{Create measurement grid}
In the measurement grid the cell probabilities for being free $P_\text{\emph{free}}$  and static occupied $P_\text{\emph{static}}$ based on the latest measurements are stored. All updates for the states are first calculated independently and are normalized jointly in a later step. In our implementation we calculate free space with a Gaussian probability density function (PDF) $f_d(d)$ with $\mu_{f} = 0$ and a standard deviation $\sigma_{\!f}$ depending on sensor and resolution. The variable is the distance \textit{d} between the cell center $(x_c,y_c)$ and the measurement. 
\begin{equation}
    d_c = \left\|\left[\begin{matrix}x_r\\y_r\end{matrix}\right]-\left[\begin{matrix}x_c\\y_c\end{matrix}\right]\right\|_2
\end{equation}
\begin{equation}
    f_d(d) = \frac{1}{\sigma_{\!f}\sqrt{2\pi}} \exp\left( -\frac{1}{2}\left(\frac{d-\mu_{f}}{\sigma_{f}}\right)^{2}\right) \label{eq2}
\end{equation}
\begin{equation}
    P_\text{\emph{free}} =1 - f_d\left(d_c\right) 
\end{equation}

For $P_\text{\emph{static}}$ we incorporate both velocity and position of the measurement. The general idea is that the closer a cell is to a measurement with zero-velocity the more likely it is to be static occupied. This is possible due to the velocity measurement of the radar sensor. We define a 3D-multivariate Gaussian density distribution $f_v(\mathbf{v}_r)$ with distance and velocity equal to zero as $\boldsymbol\mu_{s}$ and a covariance $\boldsymbol\Sigma_{s}$. The variable $\mathbf{x}_s$ is a combination of the distance $d_c$ and the velocity of used $\mathbf{x}_r$.
\begin{equation}
    \mathbf{x_s} = \left[d_c,v_{r,x},v_{r,y}\right]^T
\end{equation}
\begin{equation}
    P_\text{\emph{{static}}} =\frac{1}{|\sqrt{2\pi\boldsymbol\Sigma_{s}}|} \exp\left(\! -\tfrac{1}{2}\left(\mathbf{x_s}\!-\!\boldsymbol\mu_{s}\right)^{T}\mathbf{\Sigma}_{s}^{-1}\left(\mathbf{x_s}\!-\!\boldsymbol\mu_{s}\right)\!\right) \label{eq4}
\end{equation}

Including the velocity into the formulation allows to better estimate if a cell is static compared to only the distance to a measurement. For both cases, the cells that need to be updated can be determined using ray tracing algorithms and it's variations. Each measurement is then used to update the existing occupancy grid following a BOF formulation.

\subsubsection{Update particle weights}
Parallel to creating the measurement grid the weights of the particles need to be determined. In a BOF formulation the existing weight $w_{t-1}$ with new information based on the latest measurement. The weights determine which particle are kept during the resampling step of the particle filter. Each particle is defined as follows:
\begin{equation}
    \mathbf{x}_p = \left[x_p, y_p, v_{p,x}, v_{p,y}, w_t\right]^T
\end{equation}
For the weight update two approaches based on the formulation in \cite{b8} can be utilized. In the first one the weight is calculated relative to the distance $d_{p}$ between particle and nearest measurement (nearest neighbour). A decay factor $\epsilon$ is used to model the increasing uncertainty in the previous weight.
\begin{equation}
    w_{position} = \underbrace{f_d(d_{p})}_{update}\underbrace{(1-\epsilon)w_{t-1}}_{prior}
\end{equation}
Even though the velocity is not included, the assumption is that only particles which move in the correct direction can accumulate enough weight to be consistently resampling. This approach is also used with lidar data, which don't provide direct velocity information.

The second method includes velocity in a way similar to the static cell computation. Particles which have a velocity similar as the closest measurement receive a high weight. In addition a proximity factor, based on the distance to the measurement, is used to ensure only particles within reasonable distance are rewarded. In case there is no measurement, the previous weight is kept.  A corresponding update term is defined as follows:
\begin{equation}
    \mathbf{v}_p = (v_{p,x},v_{p,y})
\end{equation}
\begin{equation}
    d_p= \left\|\left[\begin{matrix}x_r\\y_r\end{matrix}\right]-\left[\begin{matrix}x_p\\y_p\end{matrix}\right]\right\|_2
\end{equation}
\begin{equation}
    w_{velocity} = \underbrace{f_d(d_{p})f_v(\mathbf{v}_p)}_{update}\underbrace{(1-f_d(d_{p}))(1-\epsilon)w_{t-1}}_{prior} \label{eq5}
\end{equation}
The function $f_v(\mathbf{v}_p)$ is a PDF similar to (\ref{eq4}) with $ \mu_v\!=\!(v_{r,x},v_{r,y})$ and a respective $\mathbf{\Sigma}_v$. An discussion on the different weight calculations and the impact on performance when choosing either is provided in section \ref{section_4}.
\subsubsection{Update the dynamic occupancy grid}
Lastly an important part of the dynamic occupancy grid framework is the joint update of all states. In a first step a normalization factor $q_\text{\emph{norm}}$ is calculated per cell by suming up the probabilities of all states ($P_{free},P_{static}$) and weight $w_k$ of each single particle in the respective cell:
\begin{equation}
    q_\text{\emph{norm}} = P_\text{\emph{free}} + P_\text{\emph{{static}}} + \sum_k{w_k}
\end{equation}
Using $q_{norm}$ each of the final cell states  can be calculated:
\begin{equation}
    P(emp) = \frac{P_\text{\emph{free}}}{q_\text{\emph{norm}}}
\end{equation}
\begin{equation}
    P(occ \cap v=0) = \frac{P_\text{\emph{static}}}{q_\text{\emph{norm}}}
\end{equation}
\begin{equation}
    P(occ \cap v\neq0) = \frac{\sum_k{w_k}}{q_\text{\emph{norm}}}
\end{equation}

\section{Dual Weight Particle Filter}
In this section the challenge of measuring the velocity of objects moving lateral to the sensor using only radar is described and we propose a dual-weight particle filter in order to compensate this shortcoming.

\subsection{Challenges in Radar-Only Setups}
\begin{figure}[htbp]
\centerline{\includegraphics[width=\linewidth]{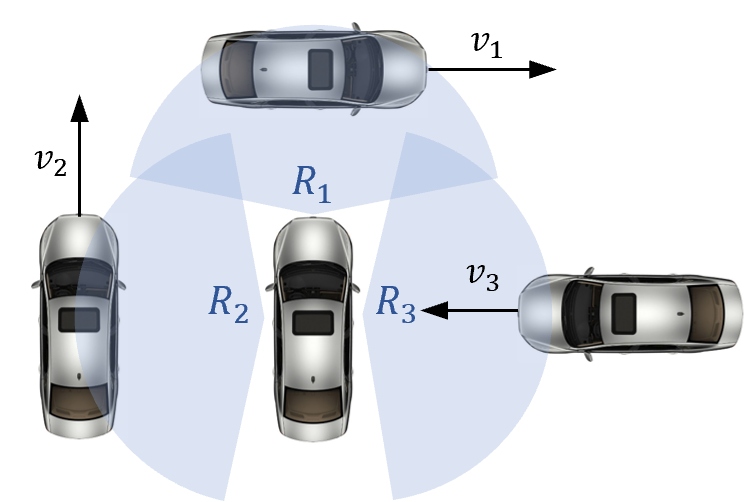}}
\caption{Examples for when the radial velocity $\Dot{R}$, with $R$ denoting the radius, is equivalent to the object velocity, and when it is seen as non moving: $\Dot{R_1}=0$ (but $v_1 \neq 0$), $\Dot{R_2} = 0$ (but $v_2 \neq 0$), $\Dot{R_3} = v_3$.} 
\label{radar}
\end{figure}

A characteristic of radar sensors is the ability to directly measure the radial velocity using the Doppler effect. In many situations already having this partial velocity measurement is already highly beneficial. As shown in \cite{b8} a radar and lidar dynamic occupancy grid map outperforms the lidar-only version. However, when an object is moving lateral (tangential) to the sensor, the radial distance is not changing. Hence, the object is measured with a radial velocity of zero and seen as non-moving (Fig.~\ref{radar}). For the task of dynamic occupancy grid mapping, where we want to distinguish between static and dynamic objects, this can cause problems when using radar as the only sensor. An obvious solution would be to use the position of the measurement only as in (\ref{eq4}). But this would require to significantly increase the number of particles in order to improve the chance of getting the correct velocity and thereby significantly add computation load.  In addition it would mean to completely neglect the benefits radar can provide in terms of velocity estimation. Hence in in section \ref{section_dual_weights} we propose a dual weight approach to optimally use radar and achieve good results with only a small number of particles.

\subsection{Dual Weights}\label{section_dual_weights}
\begin{figure}[htbp]
\centerline{\includegraphics[width=\linewidth]{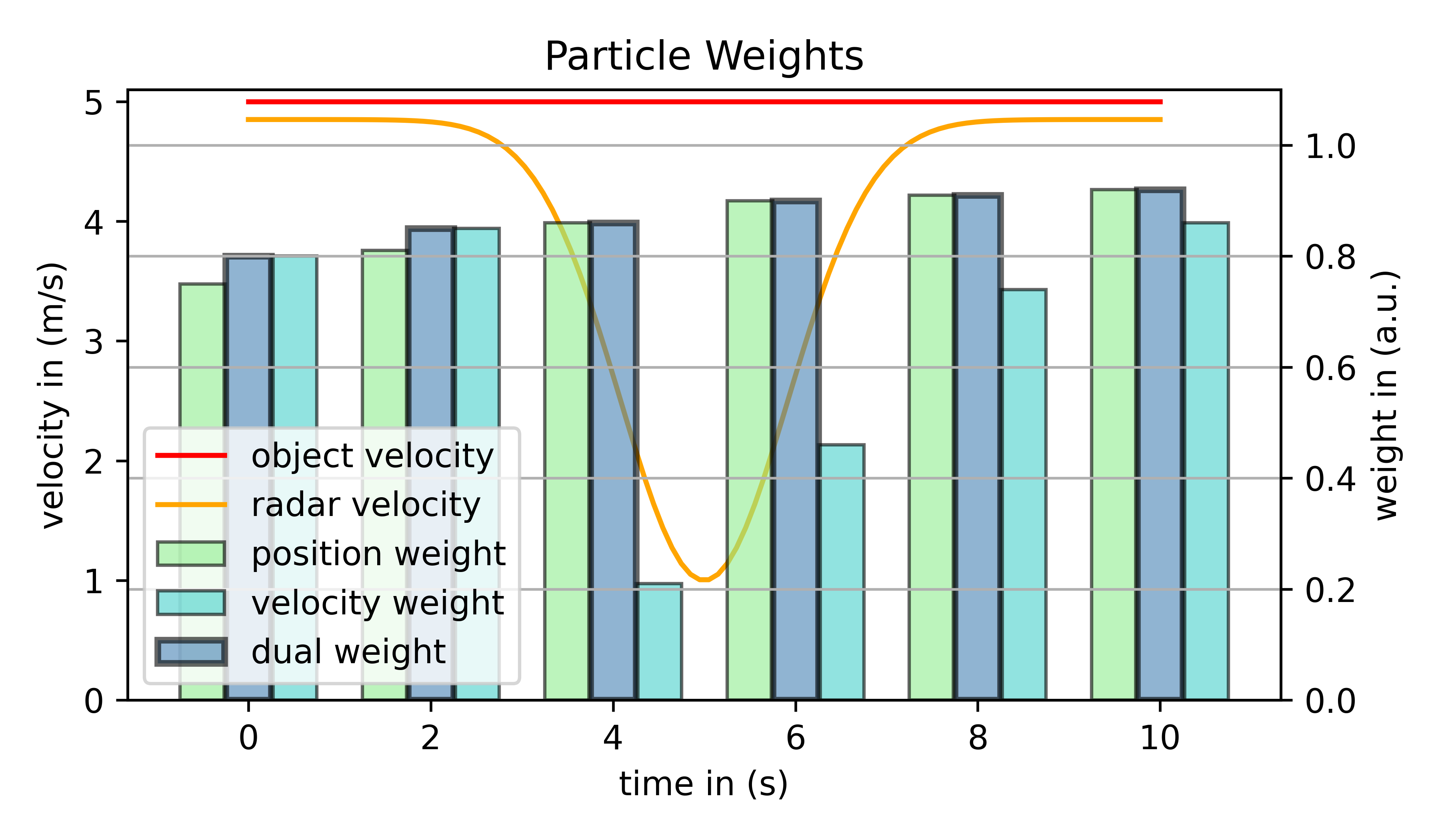}}
\caption{Dual Particle Weights: When the measured radial velocity is close to the objects velocity (0s-2s,8s-10s) the velocity weight and position weight behave similar. When the radial velocity differs from the objects (3s to 8s), the particle can still keep a high weight.} 
\label{dual_weight behavior}
\end{figure}

The main factor in determining if a cell is dynamic or static is the correct weight calculation of the particle weights. In (\ref{eq4}) and (\ref{eq5}) we introduced how to calculate the weights $w_{position}$ and $w_{velocity}$. Both of them, with their own set of advantages and shortcomings. Neither of weight calculations alone is sufficient to create accurate dynamic grid maps.Therefore, we propose a simple adaption to the framework by using two weights instead of one and continuously update both of them. The reasoning behind is that in situations where one of the weight calculations is sub optimal or wrong, the other one compensates (Fig.~\ref{dual_weight behavior}). Hence, improving the particle filter estimation and consequently overall dynamic grid performance. The dual weight approach only requires minor changes in the algorithm.
\begin{itemize}
    \item Extend the particle state to have two weights:
    \begin{equation}
        \mathbf{x}_p = \left[x_p, y_p, v_{p,x}, v_{p,y}, w_{position},w_{velocity}\right]^T
    \end{equation}
    \item Each iteration update and consistently track both weights $w_{position}$ and $w_{velocity}$.
    \item Determine the weight for the re-sampling step using a max operator:
    \begin{equation}
        w_{resample} = \max{(w_{position},w_{velocity})}
    \end{equation}
\end{itemize}
In Fig.~\ref{dual_weight behavior} an simplified example of how the introduced weights change in different situations. The assumption is that the particle is moving with a velocity similar to that of the object.

\section{Evaluation}\label{section_4}
In this section we evaluate the performance of the different single weight solutions and show that our dual weight approach outperforms either of them. The evaluation is done with simulated data generated using the MATLAB Radar and Autonomous Driving Toolboxes. As a sensor setup four short-range radars (SSR) at the wheel positions and one longe-range radar in the front center are used. During evaluation we focus on the ability to track moving parts in the grid map. A cell is classified as dynamic when  $P(occ \cap v\neq0)$ is larger then 0.6. Particles in dynamic cells are clustered using a DBSCAN algorithm \cite{b22} and weighted averages for position and velocity are taken. Those are then compared to the ground truth.

\begin{figure}[htbp]
\centerline{\includegraphics[width=\linewidth]{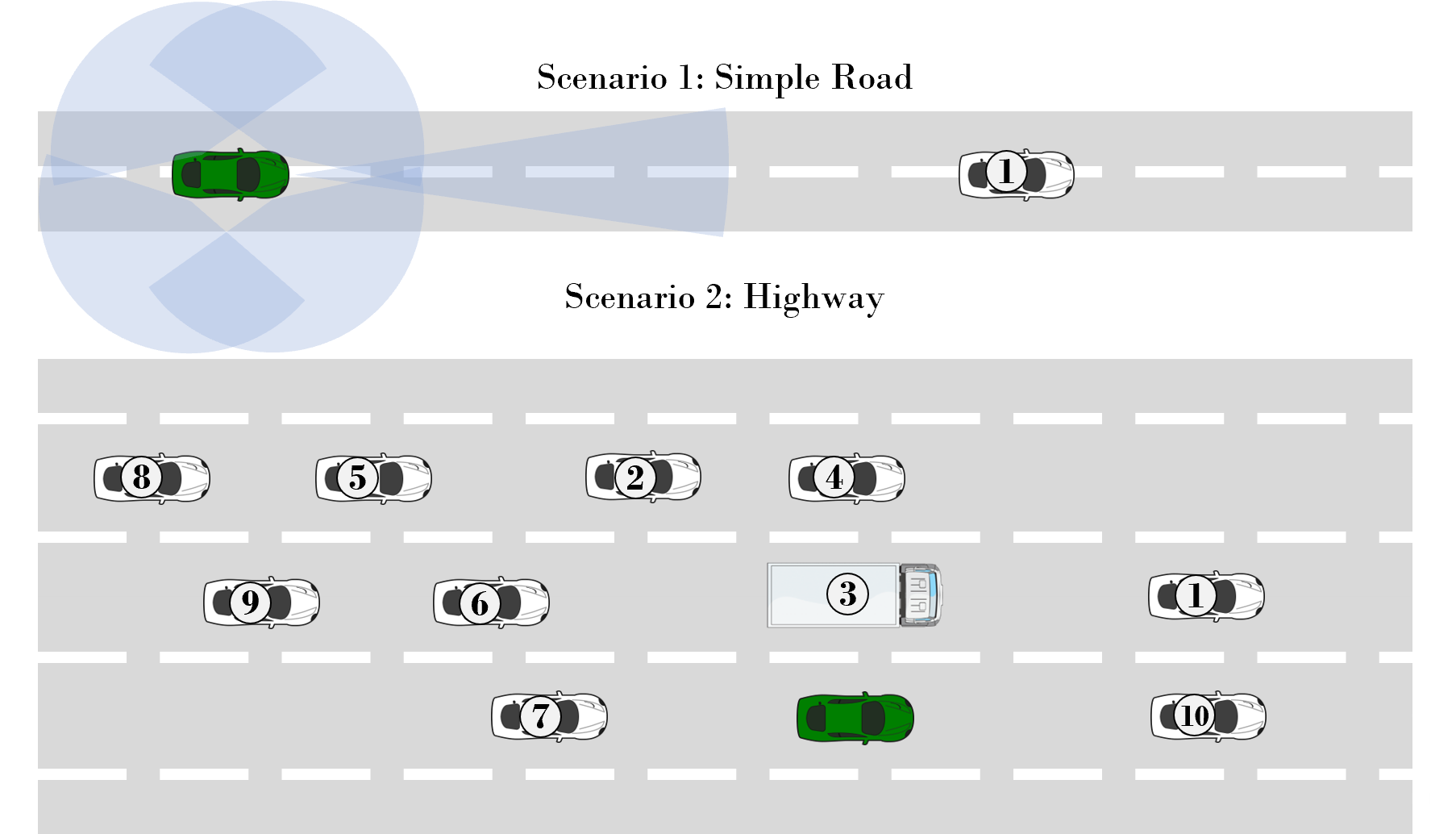}}
\caption{Overview of the testing scenarios used for evaluating the presented approach. The green car represents the ego-vehicle.} 
\label{scenarios}
\end{figure}

We evaluate the object tracking and dynamic estimation on two scenarios, simple following maneuver (simple road) and a more complex highway scenario. In both cases, vehicles move between 70 km/h and 130 km/h. The map size is 200m x 25m with a resolution of 0.5 and 10000 particles. Fig.~\ref{scenarios} shows the structure of the testing scenarios. 

\subsection{Scenario 1: Simple Road}
The simple road scenario is used to showcase how the different weights calculations affect the ability of the particle filter to accurately determine velocity and position of an object. In this scenario the measured radial velocity is close to the object velocity. As metrics we use the following: $\Delta x$ and $\Delta v$ are the average norm of difference between cluster value and ground truth object. As a measure for how fast the filter converges we take the time $t_d$ when the first cluster can be found consistently. Furthermore we compare how much of the whole scenario has been tracked by calculating the average duration $D$ an object is tracked relative to the total time. The results are summarized in table \ref{tab1}.

\begin{table}[htbp]
\caption{Results for Scenario 1}
\begin{center}
\begin{tabular}{|c|c|c|c|}
\hline
\textbf{Evaluation}&\multicolumn{3}{|c|}{\textbf{Particle weights}} \\
\cline{2-4} 
\textbf{metric} & \textbf{\textit{position weight}}& \textbf{\textit{velocity weight}}& \textbf{\textit{dual weight}} \\
\hline
$\Delta x$& 1.20 m& 0.84 m & \textbf{0.53} m  \\
$\Delta v$& 0.55 m/s&\textbf{0.22 m/s} & 0.39 m/s \\
$t_d$& 0.92s& \textbf{0.29s} & 0.49s  \\
$D$& 85 \% &89 \% & \textbf{91} \%  \\
\hline
\end{tabular}
\label{tab1}
\end{center}
\end{table}

It shown that when velocity is included in the weight computation (velocity/dual) outperform the only position based version, both in time to converge and accuracy. This shows that using the radial velocity when it's correct boosts performance. The differences between dual weight and velocity weight can be explained with the randomness in the particles. Although the velocity weight should deliver a better estimate there is still always the chance that a particle appears directly next to a measurement and is rewarded with a high position weight. This in return affects the overall performance of the dual solution as the correct velocity is not guaranteed.  
\subsection{Scenario 2: Highway}
In this case we want to prove that the dual solution is able to compensate for lack of tangential velocity measurement from a radar sensor. The same metrics as in the previous scenario are used except the time until the first clustered object. Results are summarized in table \ref{tab2}. %We replace this with the average number of objects per scan.

\begin{table}[htbp]
\caption{Results for Scenario 2}
\begin{center}
\begin{tabular}{|c|c|c|c|}
\hline
\textbf{Evaluation}&\multicolumn{3}{|c|}{\textbf{Particle weights}} \\
\cline{2-4} 
\textbf{metric} & \textbf{\textit{position weight}}& \textbf{\textit{velocity weight}}& \textbf{\textit{dual weight}} \\
\hline
$\Delta x$& 2.3 m& 3.3 m & \textbf{1.8} m  \\
$\Delta v$& 3.4 m/s& 3.3 m/s & \textbf{2.8 m/s} \\
$D$& 34 \% &40 \% & \textbf{63} \%  \\
\hline
\end{tabular}
\label{tab2}
\end{center}
\end{table}

\begin{figure}[htbp]
\centerline{\includegraphics[width=\linewidth]{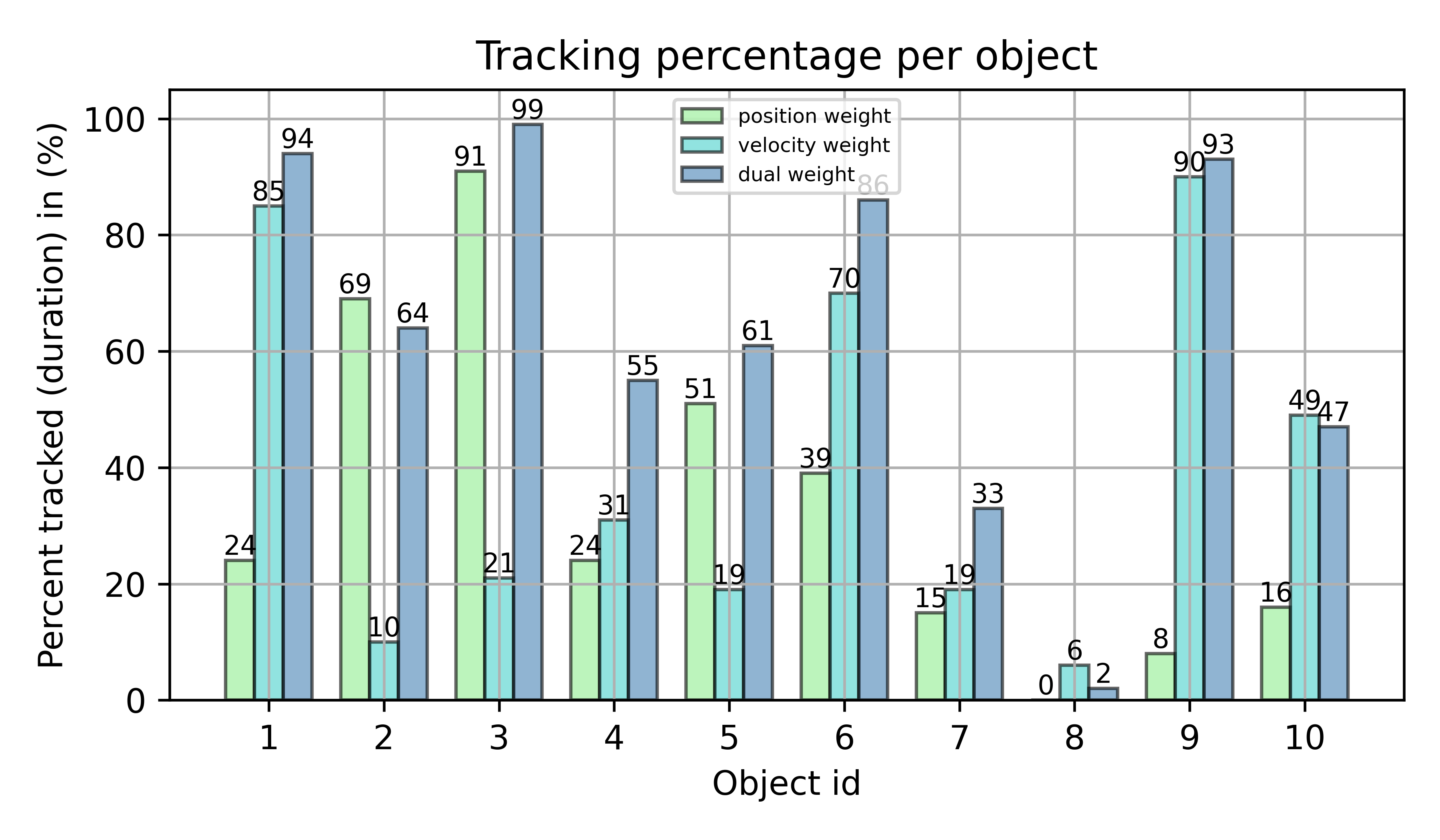}}
\caption{The results for the average tracking time per object for each of the three computations is shown. It can be observed that the dual weight has by far the longest tracking time.} 
\label{highway}
\end{figure}

In all metrics the dual weight approach can achieve the best results. The difference in performance becomes especially visible when looking at the tracking duration on an object level in Fig.~\ref{highway}.

\begin{figure}[htbp]
\centerline{\includegraphics[width=\linewidth]{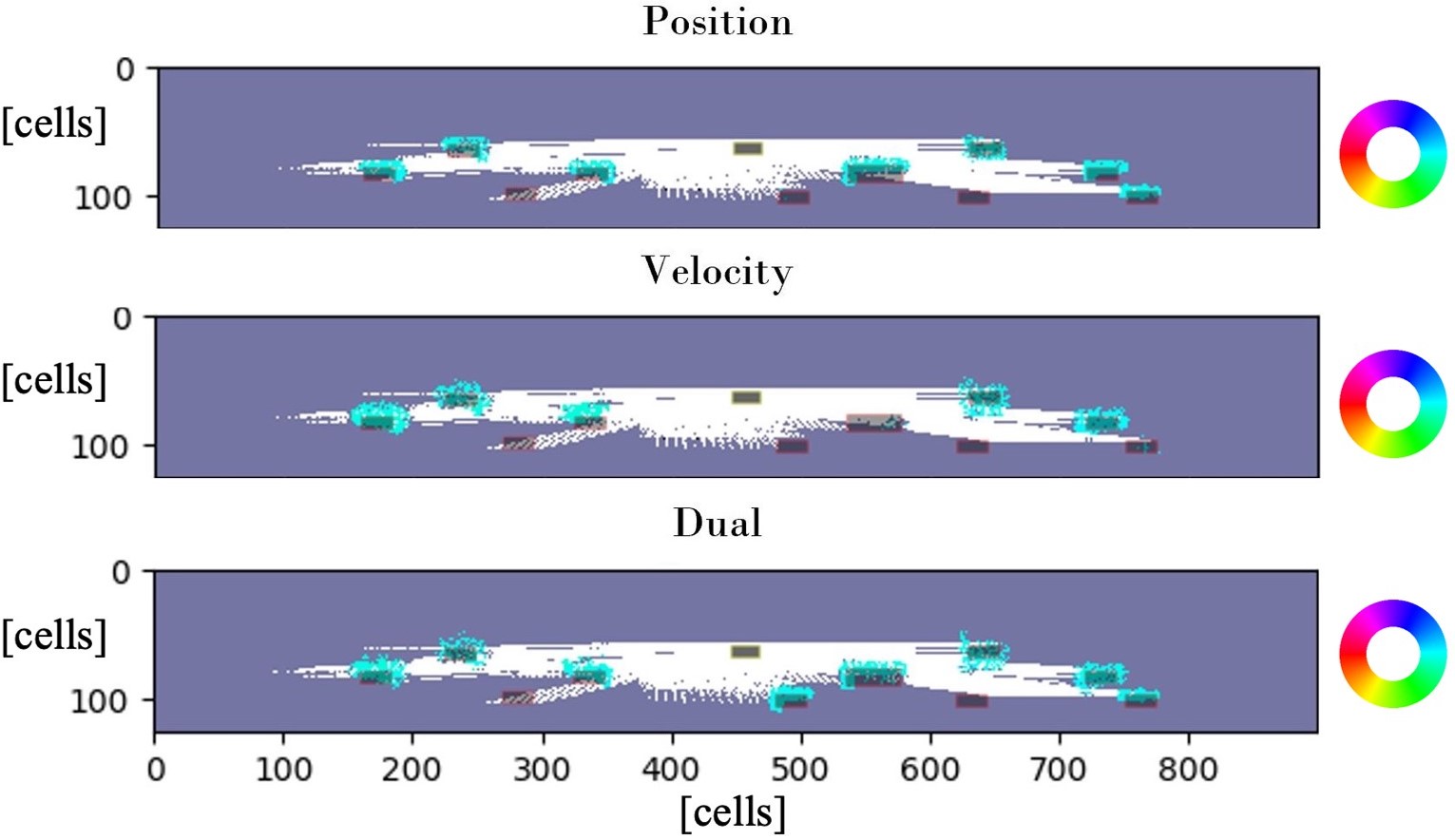}}
\caption{Examples of generated dynamic occupancy grids are given. The occupancy grid based on the dual weight provides the most complete picture. The color indicates \textit{dynamic} and the direction of the movement. A \textit{free} cell is white and \textit{static} black, although not visible in the example.} 
\label{dogma}
\end{figure}

The dual weight shows exactly the desired behavior. When one of the weights doesn't allow the particle filter to track an object, it can be compensated with the other one if it better suits the situation. "Object 3", a truck moving parallel to the ego-vehicle (Fig.~\ref{scenarios}) is a good example for this. Although clearly visible for the sensor, the measured radial velocity is significantly lower than the actual one, which makes it difficult for the velocity weight to track it. On the other side the position weight has no such issues at it's only relying on the position. Our introduced dual weight approach utilizing the weights in parallel, can outperform each of the single weights. The described behavior is also directly visible in the created dynamic grids in Fig.~\ref{dogma}.

\section{Conclusion and Future Work}\label{section_5}
A dual-weight particle filter as an extension for the Hybrid Sampling Bayesian Occupancy Filter has been introduced. It enables the framework to reliably work with only radar data. During the evaluation with simulated data it has been shown that the method can overcome common short comings of radar sensors and that it works better than existing single weight solution. In future work we would extend the evaluation to real data and make a thorough comparison with a lidar based solution, also in regard to free space and occupancy calculation as well as to quantify the computational load. Further, the impact of ghost objects (multi path propagation) on dynamic grids, another common issue with radar sensors, will be quantified. In a next step we would then like to use semantic information from camera to further mitigate some of the deficiencies of radar sensors.   
%\section*{Acknowledgment}
%We want to express our thanks to our industrial partners who enabled this work. We especially thank Maxime %Derome, Ola Thomson and Hossein Nemati from Huawei Technologies Sweden AB for their cooperation and fruitful %discussions as well as for their support. 

\end{document}